\documentclass[10pt,twocolumn,letterpaper]{article}

\PassOptionsToPackage{dvipsnames,table}{xcolor}
\usepackage{cvpr}              

\definecolor{cvprblue}{rgb}{0.21,0.49,0.74}

\usepackage{multirow} 
\usepackage{graphicx,makecell} 
\usepackage[accsupp]{axessibility}  
\usepackage[pagebackref,breaklinks,colorlinks,allcolors=cvprblue]{hyperref}

\def\paperID{15001} 
\def\confName{CVPR}
\def\confYear{2025}

\title{RENO: Real-Time Neural Compression for 3D LiDAR Point Clouds}

\author{Kang You\textsuperscript{1}, Tong Chen\textsuperscript{1}\thanks{Corresponding author.}~, Dandan Ding\textsuperscript{2}, M. Salman Asif\textsuperscript{3}, Zhan Ma\textsuperscript{1}\\
\textsuperscript{1}Nanjing University, \textsuperscript{2}Hangzhou Normal University, \textsuperscript{3}University of California Riverside\\
{\tt\small youkang@smail.nju.edu.cn, chentong@nju.edu.cn, dandanding@hznu.edu.cn,}\\
{\tt\small sasif@ucr.edu, mazhan@nju.edu.cn}
}

\begin{document}
\maketitle
\begin{abstract}
Despite the substantial advancements demonstrated by learning-based neural models in the LiDAR Point Cloud Compression (LPCC) task, realizing real-time compression—an indispensable criterion for numerous industrial applications—remains a formidable challenge. This paper proposes RENO, the first real-time neural codec for 3D LiDAR point clouds, achieving superior performance with a lightweight model. RENO skips the octree construction and directly builds upon the multiscale sparse tensor representation. Instead of the multi-stage inferring, RENO devises sparse occupancy codes, which exploit cross-scale correlation and derive voxels' occupancy in a one-shot manner, greatly saving processing time. 
Experimental results demonstrate that the proposed RENO achieves real-time coding speed, 10 fps at 14-bit depth on a desktop platform (e.g., one RTX 3090 GPU) for both encoding and decoding processes, while providing 12.25\% and 48.34\% bit-rate savings compared to G-PCCv23 and Draco, respectively, at a similar quality. RENO model size is merely 1MB, making it attractive for practical applications. The source code is available at \url{https://github.com/NJUVISION/RENO}.
\end{abstract}
\section{Introduction}
\label{sec:intro}

\begin{figure}[t]
\centering
\includegraphics[width=\linewidth]{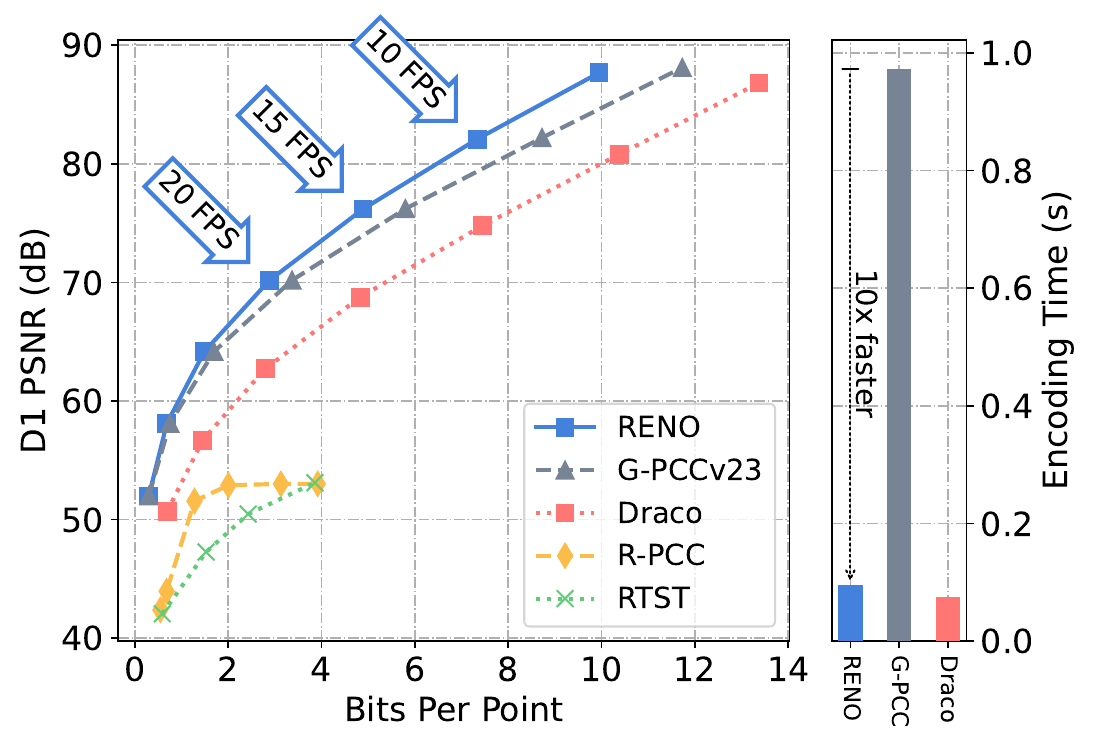}
\caption{Left: Rate distortion performance comparison on the KITTI dataset. Right: Encoding time comparison for 14 bit ($\pm$ 8 mm precision) LiDAR scan, where RENO operates 10 $\times$ faster than the latest G-PCCv23 standard, achieving a runtime of 10 frames per second. Notably, the encoding time encompasses all durations including preprocessing, network inference, and arithmetic coding. Our decoding time is comparable to the encoding.}
\label{fig:teaser}
\end{figure}


\begin{figure*}[t]
\centering
\includegraphics[width=\textwidth]{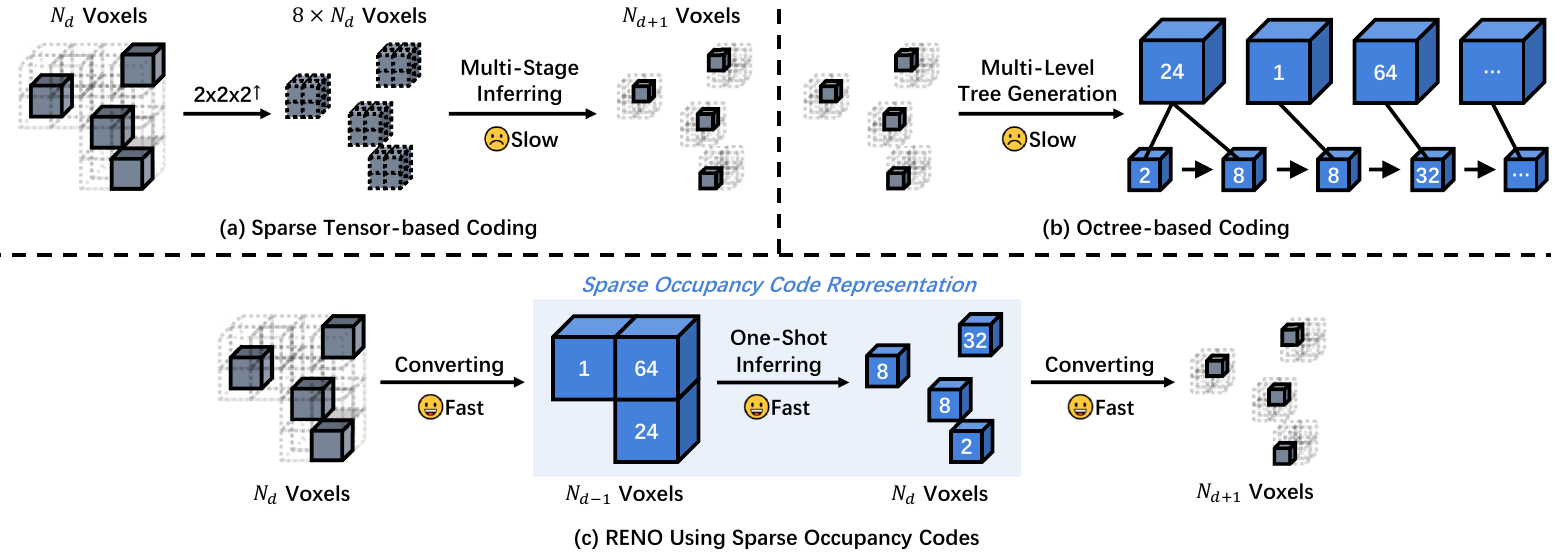}
\caption{Comparison of learned LPCC pipelines. The refinement of point clouds from depth (or scale)  $d$ (with $N_d$ voxels) to depth $d+1$ (with $N_{d+1}$ voxels) is employed as a toy example for better illustration. (a) Current sparse tensor-based methods predominantly employ a multi-stage inferring pipeline. However, the involved upsampling operation introduces $8 \times N_d$ voxels for neural network-based inference, leading to significant computational costs. (b) The octree-based method leverages the tree representation to obtain extensive contextual information but requires a time-intensive process for multi-level octree generation. (c) RENO introduces sparse occupancy codes to avoid multi-level tree generation and facilitate high-speed one-shot inferring, delivering real-time compression.}
\label{fig:intro}
\end{figure*}




LiDAR point clouds serve as essential representations of 3D environments and are widely utilized in robotics~\cite{yang2024lisa,wang2022efficient}, autonomous driving~\cite{hao2024mapdistill,2024chen,mao20233d}, and 3D mapping~\cite{zhong20243d,2023MiaoMapping}. In time-critical applications, such as autonomous driving, the continuous influx of point cloud data necessitates advanced and real-time\footnote{Such a ``real-time'' criteria is defined by the frequency to collect LiDAR data, which is typically set to 10 Hz.} LiDAR Point Cloud Compression (LPCC) techniques 
to effectively facilitate on-device storage~\cite{abbasi2022lidar}, network  transmission~\cite{liu2022communication}, and collaborative decision-making~\cite{cui2022coopernaut} within dynamic environments. 


Conventional rules-based LPCC methods face significant challenges in balancing rate-distortion performance and low compression latency. For example, the Moving Picture Experts Group (MPEG) committee has introduced the G-PCC standard~\cite{li2024mpeg,GPCCdescription}, which is implemented in the TMC13 software, demonstrating convincing compression performance but currently lacks real-time capabilities~\cite{santos2024coding}. In contrast, Google's Draco~\cite{Draco} offers low latency but exhibits inferior rate-distortion efficiency. Alternatively, range image-based methods~\cite{stathoulopoulos2024recnet,R-PCC,RTST,zhao2022real,heo2022flicr,zhou2022riddle,nenci2014effective} suffer from projection errors, which result in a loss of information~\cite{wang2023multi} and may hinder the performance of reconstruction-related tasks~\cite{wu2021detailed}.

In recent years, neural codecs~\cite{Unicorn_Part1,SparsePCGC,ECM_OPCC,EHEM,OctAttention,OctSqueeze,SparseContextNet,Pointsoup,xue2024neri} have garnered significant attention. While these methods present impressive rate-distortion performance, their space-time complexity makes them unusable for real-time LiDAR data compression and transmission. For example, one of the current state-of-the-art methods, Unicorn~\cite{Unicorn_Part1}, demonstrates favorable complexity; however, it still takes approximately 2 seconds to encode one 14-bit LiDAR frame on an RTX 3090 GPU. This prompts the question: ``\emph{Why do current neural codecs face runtime bottlenecks with 3D LiDAR point clouds, and how can we overcome these issues?}''



In multiscale sparse tensor-based solutions~\cite{SparsePCGC,Unicorn_Part1}, the compression of current-scale voxels\footnote{Voxels or points refer to occupied coordinates in a point cloud frame.} involves the dyadic upscaling of (causally-available) lower-scale voxels to form proper contexts for entropy coding. Although Unicorn~\cite{Unicorn_Part1} grouped current-scale voxels to perform multi-stage computation to remove autoregressive processing, real-time compression remains challenging (see Fig.~\ref{fig:intro}(a)). On the other hand, although octree models~\cite{GPCCdescription,OctSqueeze} can use a node's occupancy symbol to infer the occupancy status of its eight child nodes, the process of constructing the octree of each input point cloud is notably complex (see Fig.~\ref{fig:intro}(b)). Some recent explorations~\cite{SparseContextNet,fan2023multiscale} attempted to improve computational efficiency by integrating octree and sparse tensor representations, but they did not radically solve the problem because of 1) pre-generating octree structure and 2) multi-stage processing using upscaled voxels (from lower depth).






 We introduce RENO that employs the multiscale sparse tensor representation to skip time-consuming octree generation in Fig.~\ref{fig:intro}(b) and proposes to compress 8-bit sparse occupancy codes in a one-shot manner at each scale (or depth) to avoid multi-stage processing as in Fig.~\ref{fig:intro}(a). 
 As a result, the compression of LiDAR point cloud geometry is formulated as the compression of sparse occupancy code sequence scale by scale. 
 
 Sparse occupancy codes can be directly derived using fixed-weights sparse convolution along with dyadic downscaling when generating multiscale sparse tensors. They carry the same categorical values (1–255) as the occupancy symbols in octree-structured nodes, which are used to explicitly indicate occupied child nodes in octree representation (i.e., now dyadically upscaled voxels at the following high scale), but they are \emph{orderless} instead of tree-ordered in an octree model. 
 Then, they are compressed with contexts conditioned on the available information (e.g., occupancy codes, occupied coordinates) from the lower scale.  

The proposed RENO makes the following contributions:

\begin{itemize}
    \item It is probably \emph{the first} neural model to compress 3D LiDAR point cloud geometry in a real-time manner, which acts in concert with the spinning speed of popular LiDAR sensors.

    \item It also outperforms the latest LiDAR compression standard G-PCC, Draco, etc, owing to the cross-scale context modeling to compress the proposed sparse occupancy codes scale by scale.


    \item It is also a lightweight model with a size of merely 1 MB, making it attractive to industrial practitioners. 

\end{itemize}

\begin{figure*}[t]
\centering
\includegraphics[width=\textwidth]{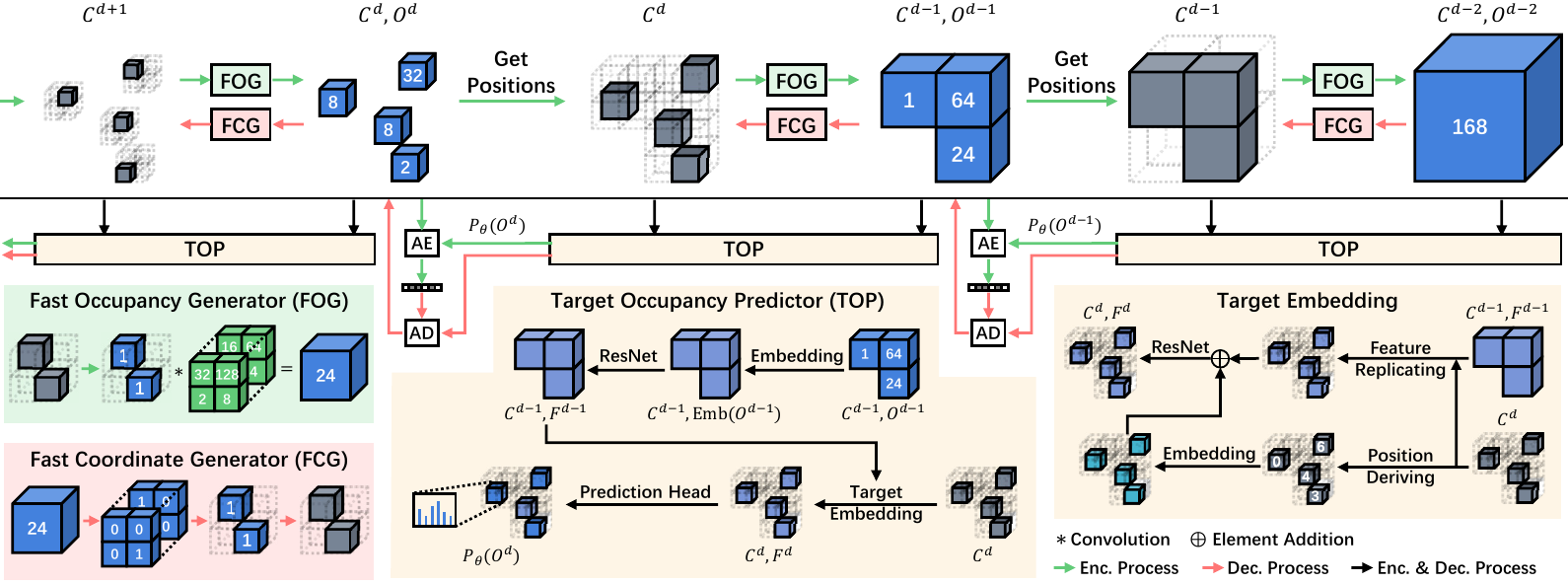}
\caption{RENO. We identify that the real-time bottleneck of current neural codecs resides in two stages: preprocessing and neural inference.  To surmount these obstacles, this approach endeavors to optimize efficiency by (i) minimizing preprocessing delays through the efficient acquisition of occupancy codes directly within sparse space via the developed Fast Occupancy Generator (FOG) and Fast Coordinate Generator (FCG); (ii) optimizing neural inference by efficiently embedding features from sparse occupancy codes to the next-level target positions, which prompted the development of the Target Occupancy Predictor (TOP).}
\label{fig:framework}
\end{figure*}

\section{Related Works}
\label{sec:relwork}

Numerous studies have advanced the field of LiDAR Point Cloud Compression (LPCC). To appreciate the inherent sparsity of LiDAR point clouds, a given input frame has to be geometrically structured into trees, multiscale sparse tensors, or range images for context modeling.

{\bf Tree}.  One is the simple {\it kd-tree}~\cite{pacheco2021point}, which has been used in Google's Draco~\cite{Draco} for real-time compression.  Another popular one is the {\it octree}, which was widely studied~\cite{wang2024optimized,anand2019real,rusu2011PCL} and incorporated into the MPEG G-PCC standard~\cite{GPCCdescription}.  G-PCC adopted heuristic rules to mine neighbors to form contexts for entropy coding, showing improved compression efficiency but a much slower processing speed than Draco.


More recently, OctSqueeze~\cite{OctSqueeze} and OctAttention~\cite{OctAttention} pioneered efforts by introducing learnable contextual models. Such a learning-based context modeling was subsequently advanced by the development of ECM-OPCC~\cite{ECM_OPCC}, EHEM~\cite{EHEM}, and others~\cite{cui2024octree,cui2023octformer}, yielding noticeable progress of coding efficiency. Yet, generating an octree-structured point cloud for context modeling creates a notable runtime bottleneck.

{\bf Multiscale Sparse Tensor.} Besides the octree representation, the multiscale sparse tensor formulation was introduced in SparsePCGC~\cite{SparsePCGC} and has been further refined in subsequent research endeavors (e.g., SparsePCGCv2~\cite{xue2022efficient}, Unicorn~\cite{Unicorn_Part1}, etc.). However, pruning upsampled sub-voxels to effectively determine their occupancy statuses often relies on multi-stage processing, which is also time-consuming.

While certain approaches~\cite{SparseContextNet,fan2023multiscale} attempt to enhance computational efficiency by integrating octree-based occupancy codes with sparse tensor representations, these methods remain fundamentally limited by the abovementioned constraints.



{\bf Range Images.} Alternatively, considering laser sensor's physical properties, the LiDAR point cloud frame can be projected to produce a 2D range image, which has consequently inspired the development of various range image-based compression methods~\cite{stathoulopoulos2024recnet,R-PCC,RTST,zhao2022real,heo2022flicr,zhou2022riddle}. Nonetheless, projection-based approaches introduce significant degradation~\cite{wu2021detailed} and fail to provide high-fidelity reconstruction of point clouds.



\section{Method}
\label{sec:method}


As shown in \cref{fig:framework}, RENO is built upon the multiscale sparse tensor representation~\cite{SparsePCGC,Unicorn_Part1}, in which scale-wise sparse tensors of point cloud geometry (occupied coordinates, or voxels), e.g., $\{\ldots, C^{d}, C^{d-1}, \ldots\}$, are progressively generated via dyadically downscaling\footnote{Dyadic down-/up-scaling is more or less the same as the octree squeezing/expanding~\cite{SparsePCGC}. } from one scale depth to a lower one (e.g., $d$ to $d-1$). Such a representation enables voxels to be processed directly in native Cartesian space, thereby avoiding the complexities associated with tree construction. A detailed illustration of the multiscale sparse tensor can be found in~\cite{SparsePCGC}.

RENO further introduces the sparse occupancy code to strike for more efficiency. It is defined through the following state transitions:
\begin{align}
    C^{d} &=\left(C^{d-1},O^{d-1}\right) \nonumber\\
    & =\left(\left\{ c_{i}^{d-1} \right\}_{i=1}^{N_{d-1}},\left\{o_{i}^{d-1}\right\}_{i=1}^{N_{d-1}}\right),
    \label{eq:c_o_transition}
\end{align}
where $O^{d-1}=\left\{o_{i}^{d-1}\right\}_{i=1}^{N_{d-1}}$ contains the corresponding sparse occupancy code $o_{i}^{d-1} \in [1, 255]$ for each coordinates $c_{i}^{d-1}$. $N_{d-1}$ stands for the number of voxels at depth scale $d-1$.

As for the transformation defined \cref{eq:c_o_transition}, we have
\begin{itemize}
    \item $C^{d} \rightarrow \left(C^{d-1}, O^{d-1}\right)$ is to generate sparse occupancy codes $O^{d-1}$ of $C^d$, which is fulfilled using Fast Occupancy Generator (FOG) along with the dyadic downscaling;
    \item  $\left(C^{d-1}, O^{d-1}\right) \rightarrow C^{d}$ is the reconstruction of coordinates $C^{d}$ at depth $d$ from the coordinates $C^{d-1}$ at depth $d-1$ and their occupancy codes $O^{d-1}$, which is implemented using Fast Coordinate Generator (FCG) along with dyadic upscaling.
\end{itemize}



In the end, the raw tensor of the input point cloud geometry  $C^{D}$ can be formulated as:
\begin{equation}
    C^{D} = (((((C^{0}, O^{0}), O^{1}), O^{2}), \dots), O^{D-1}),
\end{equation}
suggesting that $C^{D}$ can be reconstructed losslessly by the initial state $(C^{0},O^{0})$ and a sequence of sparse occupancy codes $\mathcal{O}=(O^{1},O^{2},\dots,O^{D-1})$ at different depth scales. 

Thus, the compression of $C^{D}$ is to compress the sparse occupancy code sequence $\mathcal{O}$. Specifically, we use a neural network - Target Occupancy Predictor (TOP) to form the probability mass function $P_{\theta}(\mathcal{O})$:
\begin{align}
    P_{\theta}( \mathcal{O} ) &= \prod_{d=1}^{D-1} P_{\theta} \left( O^{d} | O^{<d}, C^{0} \right) \nonumber\\
    & = \prod_{d=1}^{D-1} P_{\theta} \left( O^{d} | O^{d-1},C^{d-1} \right), \label{eq:cond_prob}
\end{align} 

Relevant tensors $O^{d-1}$ and $ C^{d-1}$ are causally available priors from lower scale at depth $d-1$ in multiscale sparse representation framework, which are combined to exploit cross-scale correlations for context modeling. And, recalling the definition in \eqref{eq:c_o_transition}, geometry coordinates at depth $d$, e.g., $C^d$, can be easily inferred using $O^{d-1}$ and $ C^{d-1}$. As a result, in our implementation (see \eqref{eq:top}) to realize \eqref{eq:cond_prob} using TOP, we explicitly embedded as the prior to compress $O^d$.

Then, given the true distribution $P(\mathcal{O})$, the optimization for the compression can be formulated using cross-entropy:
\begin{equation}
    {\theta}^{\ast} \leftarrow \underset{\theta}{\arg\min} \mathbb{E}_{\mathcal{O} \sim P(\mathcal{O})} \left[-\log{P_{\theta}(\mathcal{O} )} \right]. \label{eq:cross_entropy}
\end{equation}

\subsection{Fast Occupancy Converters}

In previous studies~\cite{SCP,ECM_OPCC,EHEM,OctAttention,OctSqueeze}, octree-structured occupancy codes or symbols are derived by constructing the tree, which is time-consuming. Instead, we propose a Fast Occupancy Generator (FOG) to produce occupancy codes\footnote{We call them {\it Sparse Occupancy Codes} for convenience.} directly upon scale-wise sparse tensor of geometry coordinates, i.e., $\left( C^{d-1}, O^{d-1} \right) = \mathrm{FOG} \left( C^{d} \right)$.  Correspondingly, the Fast Coordinate Generator (FCG) is developed to convert sparse occupancy codes back into geometry coordinates, i.e., $ C^{d} = \mathrm{FCG} \left( C^{d-1}, O^{d-1} \right)$. 
    
   

\textbf{Fast Occupancy Generator (FOG)} is implemented using a sparse convolution with a kernel size of 2 and a stride of 2. Kernel's weights are manually fixed using $[1,2,4,8,16,32,64,128]$. 

\begin{figure}[t]
    \centering
    \includegraphics[width=1.0\linewidth]{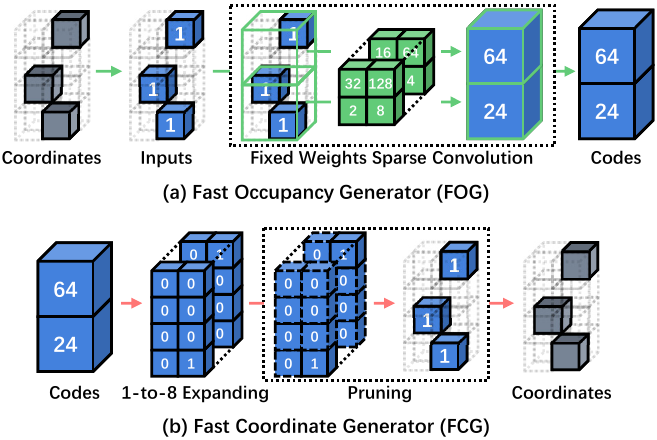}
    \caption{Illustrative implementation for Fast Occupancy Generator (FOG) and Fast Coordinate Generator (FCG).} 
    \label{fig:converters}
\end{figure}

As detailed in~\cref{fig:converters} (a), geometry coordinates of voxels are initially set to ``1'', then convoluted to produce the corresponding occupancy codes for each group of eight inter-connected nodes. 
Following the discussions above, the sparse occupancy code is an integer, ranging from 1 to 255, explicitly indicating the occupancy status of the eight nodes mentioned above. Having a code for each group of eight interconnected nodes is similar to assigning an occupancy symbol during octree construction. As seen, such a process involves dyadic downscaling or octree squeezing to merge a group of eight nodes into a single one.

In practice, fixed-weights sparse convolution easily supports parallel acceleration, offering a much lighter load for efficient computing. 



\textbf{Fast Coordinates Generator (FCG)} comprises the steps of ``expanding'' and ``pruning''. As shown in~\cref{fig:converters} (b), each sparse occupancy code is first expanded into an 8-bit binary representation, with each bit corresponding to a distinct sub-node. Such a process involves the dyadic upscaling or octree expansion to split a voxel into eight connected nodes. Following this expansion, a pruning operation is instantly conducted to just retain nodes marked with ``1'' to form the coordinate tensor.

Next, we discuss how to compress the aforementioned sparse occupancy codes efficiently.

\subsection{Target Occupancy Predictor}

A neural network-based Target Occupancy Predictor (TOP) is proposed to effectively estimate the probability of sparse occupancy codes at each scale. Thus, \eqref{eq:cond_prob} can be rewritten as 
\begin{equation}
    P_{\theta}\left(O^{d}\right) = \mathrm{TOP} \left( C^{d-1}, O^{d-1}, C^{d} \right), \label{eq:top}
\end{equation}
where $C^{d-1}$ and $O^{d-1}$ represent the coordinates and occupancy codes of sparse tensor at depth $d-1$, respectively; $C^d$ refers to the coordinate tensor at depth $d$, implying {\it target positions} of $O^{d}$; $P_{\theta}(O^{d})$ refers to the parametric probability distribution of $O^{d}$. Intuitively, the distribution can also be inferred only considering the target coordinates (i.e., $P_{\theta}\left(O^{d}\right) = \mathrm{TOP}(C^d)$). Whereas this paper further introduces $C^{d-1}$ and $O^{d-1}$ to augment prior information from a lower scale, by which cross-scale correlations can effectively improve probability estimation.

\textbf{Occupancy Feature Extraction.} Considering that $O^{d-1}$ constitute discrete values, we first utilize a naive embedding $\mathrm{Emb}$ to transform these codes into continuous vector representations, followed by a $\mathrm{ResNet}$ block that further characterizes neighborhood patterns to form aggregated features $F^{d-1} \in \mathbb{R}^{N^{d-1} \times dim} $, i.e.,
\begin{equation}
    \left( C^{d-1}, F^{d-1} \right) = \mathrm{ResNet} \left( C^{d-1}, \mathrm{Emb} \left(O^{d-1} \right) \right).
\end{equation} 
Subsequently, target embedding is devised to transfer features $F^{d-1}$ from lower scale coordinates $C^{d-1}$ to the target scale coordinates $C^d$, as illustrated in the subplot of TOP in Fig.~\ref{fig:framework}.


\textbf{Target Embedding.} As illustrated in the subplot in~\cref{fig:framework}, to enable efficient transfer of cross-scale features, the introduced target embedding consists of two steps: direct feature replication and relative position infusion. 

In the first step, features from the previous scale are directly replicated to the corresponding sub-voxels of the current scale. Let $ \tilde{F}^{d} = \{ \tilde{f}^{d}_{i} \}_{i=1}^{N_{d}} $ indicate the replicated features of the current scale, then each feature vector $ \tilde{f}^{d}_{i} $ can be expressed as follows:
\begin{equation}
\tilde{f}^{d}_{i} = f^{d-1}_{j}, \quad \forall c^{d}_{i} \in \Lambda(c^{d-1}_{j})
\label{eq:fea_replication}
\end{equation}
where $ c^{d}_{i} \in C^{d} $, $ c^{d-1}_{j} \in C^{d-1} $, and $ f^{d-1}_{j} \in F^{d-1} $; specifically, $ \Lambda(c_{j}^{d-1}) $ represents the set of coordinates corresponding to the eight sub-voxels of $c_{j}^{d-1}$.

In the second step, the relative position is identified and embedded in the replicated features. Here, we leverage the widely used octant representation to characterize the position of the sub-voxels in relation to the parent voxel. Following ~\eqref{eq:fea_replication}, the octant ${\delta}^{d}_{i} \in {\Delta}^{d}$ of each coordinate $c^d_i \in C^{d}$ can be formulated as:
\begin{equation}
    {\delta}^{d}_{i} =  {\textstyle \sum_{k=1}^{3}}  (c^{d}_{i}[k]-c^{d-1}_{j}[k])\times 2^{k-1}, \forall c^{d}_{i} \in \Lambda(c^{d-1}_{j}),
\end{equation}
where $[k]$ represents of the $k$-th coordinate element, i.e., $x$, $y$, and $z$ in this example. Subsequently, the octant is embedded to the replicated feature $\tilde{F}^{d}$, to produce the feature $F^{d}$ on the target coordinates $C^{d}$:
\begin{equation}
\begin{split}
  (C^{d}, F^{d}) &= \mathrm {ResNet} (C^d, \tilde{F}^{d} + \Delta F^{d}) \\
  &= \mathrm {ResNet} (C^d, \tilde{F}^{d} + \mathrm {Emb} ( {\Delta}^{d} )).
\end{split}
\label{eq:target_emb}
\end{equation}




\textbf{Probability Prediction.} Leveraging aggregated features $F^d$ associated with target coordinates $C^d$, a one-stage inference is performed using a multilayer perceptron ($\mathrm {MLP}$) and $\mathrm {SoftMax}$ layer to generate a 255-dimensional probability for the occupancy code in $O^{d}$, e.g.,
\begin{equation}
    P_{\theta}\left(O^{d}\right) = \mathrm {SoftMax} \left( \mathrm {MLP} \left( F^d \right) \right). \label{eq:prob_approx}
\end{equation}

Although directly predicting 8-bit codes is doable as in \eqref{eq:prob_approx}, our implementation has revealed that a bitwise two-stage strategy is preferable, not only showing an improvement of compression but also demonstrating better computational efficiency. This is because 1) predicting a 4-bit symbol is much easier than predicting an 8-bit one, not to mention that the last 4-bit symbol can also be conditionally predicted using the first 4-bit symbol; 2) transferring ($2\times N\times 16$)-element tensor of probabilities of 4-bit symbols from GPU to CPU for entropy coding significantly reduces the memory bandwidth by a factor of $\approx8$ when transferring ($N\times255$)-element tensor of probabilities for original 8-bit symbols. Here, we assume the entropy coding on the CPU as most existing works.






As for the proposed bitwise two-stage scheme,  a given 8-bit occupancy code in $O^d$ is segmented into the first 4-bit sub-codes $S^d_1$ and the last 4-bit sub-codes $S^d_2$:
\begin{equation}
P_{\theta} (O^d) = P_{\theta} (S^d_2, S^d_1) = P_{\theta} (S^d_2 | S^d_1) P_{\theta} (S^d_1).
\end{equation}
Notably, the compression process is initiated with $S^d_1$:
\begin{equation}
    P_{\theta} (S^d_1) = \mathrm {SoftMax} \left( \mathrm {MLP} \left( F^{d} \right) \right),
\end{equation}
where $F^d$ refers to the inherited features of the target coordinates, as demonstrated in~\cref{eq:target_emb}. Then, $S^d_1$ will be used as a priori knowledge for $S^d_2$ to assist its encoding, i.e., 
\begin{equation}
    P_{\theta} (S^d_2 | S^d_1) = \mathrm {SoftMax} \left( \mathrm {MLP} \left( F^{d} + \mathrm {Emb} \left(S^d_1\right)   \right) \right).
\end{equation}


\subsection{Loss Function} 

Since this paper focuses on the lossless compression of sparse occupancy codes, its loss function is to optimize the cross-entropy as in \eqref{eq:cross_entropy}, i.e.,
   $\mathcal{L} = \mathbb{E}_{\mathcal{O} \sim P (\mathcal{O})} \left [ -\log{P_{\theta}(\mathcal{O} )} \right ]$ = $\sum_{d=1}^{D-1}  \mathbb{E}_{O^d \sim P (O^d)} \left [ -\log{P_{\theta}(O^d )} \right ] $.

\section{Experiments and Discussions}

\subsection{Experimental Setup}

\subsubsection{Dataset}

\begin{table}[t]
  \centering
  \small
\caption{Quantitative BD-BR gains of RENO over other methods are reported. The encoding/decoding time is collected for 12/14 bit ($\pm$3.2cm/$\pm$8mm precision) KITTI samples. }
    \label{tab:quantitative_compare}
  \scalebox{0.86}{
      \begin{tabular}{ccrrrrr}
        \toprule
        Dataset & Metric & RTST & R-PCC & Draco & G-PCC & RENO \\
        \midrule
        \multirow{2}{*}{KITTI} 
        & D1 (\%) & -90.66 & -84.80 & -48.34 & -12.26 & - \tabularnewline
        & D2 (\%) & -70.04 & -60.17 & -48.30 & -12.23 & - \tabularnewline
        \midrule
        \multirow{2}{*}{Ford} 
        & D1 (\%) & - & -82.54 & -45.01 & -12.50 & - \tabularnewline
        & D2 (\%) & - & -68.34 & -44.97 & -12.69 & - \tabularnewline
        \midrule
        \midrule
        \multirow{2}{*}{12 bit} 
        & Enc (s) & 0.046 & 0.074 & 0.065 & 0.564 & 0.052 \tabularnewline
        & Dec (s) & 0.008 & 0.048 & 0.028 & 0.174 & 0.050 \tabularnewline
        \midrule
        \multirow{2}{*}{14 bit} 
        & Enc (s) & - & - & 0.075 & 0.973 & 0.095 \tabularnewline
        & Dec (s) & - & - & 0.032 & 0.343 & 0.090 \tabularnewline
        \bottomrule
      \end{tabular}
  }
    
\end{table}

Experiments are performed using two widely recognized datasets, namely KITTI~\cite{KITTI} and Ford~\cite{Ford}. First, compression efficiency evaluation includes rate-distortion performance, qualitative visualization, and computational complexity. Then, the downstream vision task, i.e., 3D object detection, is fulfilled to examine the effectiveness of machine visions on compressed LiDAR data.

\textbf{KITTI~\cite{KITTI}} comprises 14,999 point clouds generated by a Velodyne HDL-64E laser scanner. We adhere to the official data split, which contains 3,712 point clouds for training, 3,769 point clouds for validation, and 7,518 point clouds for testing.

\textbf{Ford~\cite{Ford}} consists of three sequences of 1,500 LiDAR scans used for MPEG standardization. Following the common test condition,  sequence 01 is used for training, while sequences 02 and 03 are reserved for evaluation.

\subsubsection{Implementation} RENO uses Python 3.9 and PyTorch 1.10 for implementation. It leverages TorchSparse++~\cite{tangandyang2023torchsparse} for efficient sparse convolution. The Adam optimizer~\cite{kingma2014adam} is employed with an initial learning rate of 5e-4 and a batch size of 1. The model is trained for 100,000 steps, with training samples quantized to 16-bit precision. All experiments are conducted on an Intel Xeon Silver 4314 CPU and one NVIDIA GeForce RTX 3090 GPU. Due to the variability in implementation details of different methods (e.g., CPU vs. GPU), the runtime comparison is intended to provide an intuitive reference of computational complexity.

\subsubsection{Benchmarking Baselines} Extensive comparisons are performed: First, since this work focuses on real-time compression, we choose three real-time compressors: Draco~\cite{Draco}, RPCC~\cite{R-PCC} and RTST~\cite{RTST}. Draco~\cite{Draco} is a tree-based solution, while RPCC~\cite{R-PCC} and RTST~\cite{RTST} uses range images to fulfill the purposes. Second, the latest (but non-real-time) G-PCCv23~\cite{GPCCdescription} standard is also included to justify the advantage of RENO's compression performance.  Additional comparison to existing learning-based LPCC like Unicorn~\cite{Unicorn_Part1}, EHEM~\cite{EHEM}, etc. will be provided in the supplemental material.



\subsection{Comparative Studies} 




\subsubsection{Rate-Distortion Performance} 
\begin{figure}[t]
    \centering
    \includegraphics[width=1.0\linewidth]{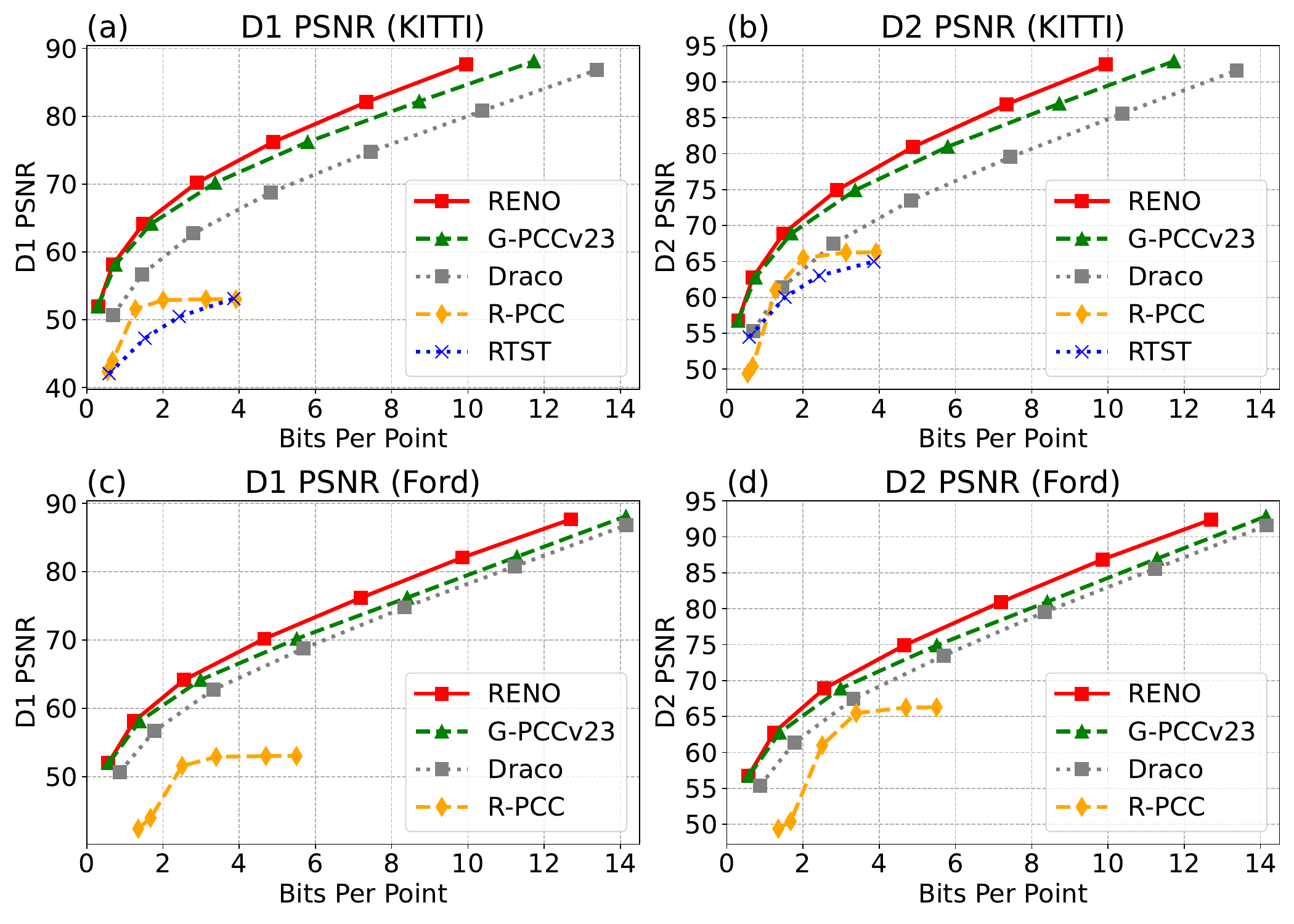}
    \caption{Rate-distortion performance comparison on KITTI (the first row) and Ford (the second row).}
    \label{fig:kitti_ford_rd}
\end{figure}

As evidenced in \cref{tab:quantitative_compare} and~\cref{fig:kitti_ford_rd}, RENO consistently outperforms the competing methods across the entire bitrate range in both point-to-point (D1) and point-to-plane (D2) metrics, showcasing its superiority.
Although G-PCCv23 achieves convincing rate-distortion performance, it lacks real-time capabilities due to the intricate rules employed for entropy coding and non-optimized reference software implementation~\cite{GPCCdescription}. On the other hand, the performances of real-time methods (e.g., Draco, R-PCC, and RTST) are noticeably inferior. Particularly, R-PCC and RTST, which adopt range image representation to facilitate low-latency compression, support only a narrow bitrate range and suffer from poor reconstruction fidelity. This limitation is due to the significant distortion introduced by the 3D to 2D projection to obtain range images~\cite{wu2021detailed}.

\subsubsection{Computational complexity} The encoding and decoding times in~\cref{tab:quantitative_compare} further support RENO’s real-time capability. For 12-bit LiDAR point clouds, RENO achieves encoding in 0.052 seconds and decoding in 0.050 seconds, closely matching or even surpassing the performance of other rules-based real-time methods. In the case of 14-bit samples, RENO continues to achieve real-time latency on average (e.g., ten frames per second), which meets the typical frequency requirements of LiDAR sensors for data collection. It is important to recognize that, owing to the specificity inherent in range image-based methods, these approaches are unable to attain standard reconstruction for a particular level of bit accuracy. Therefore, the runtimes of RPCC and RTST at 12 bits, as shown in \cref{tab:quantitative_compare}, are reported under low bit rate conditions (approximately 3 bits per point) for reference. Additionally, their runtimes at 14 bits are not applicable due to their inability to attain a comparable level of accuracy.

\begin{figure*}
    \centering
    \includegraphics[width=1.0\textwidth]{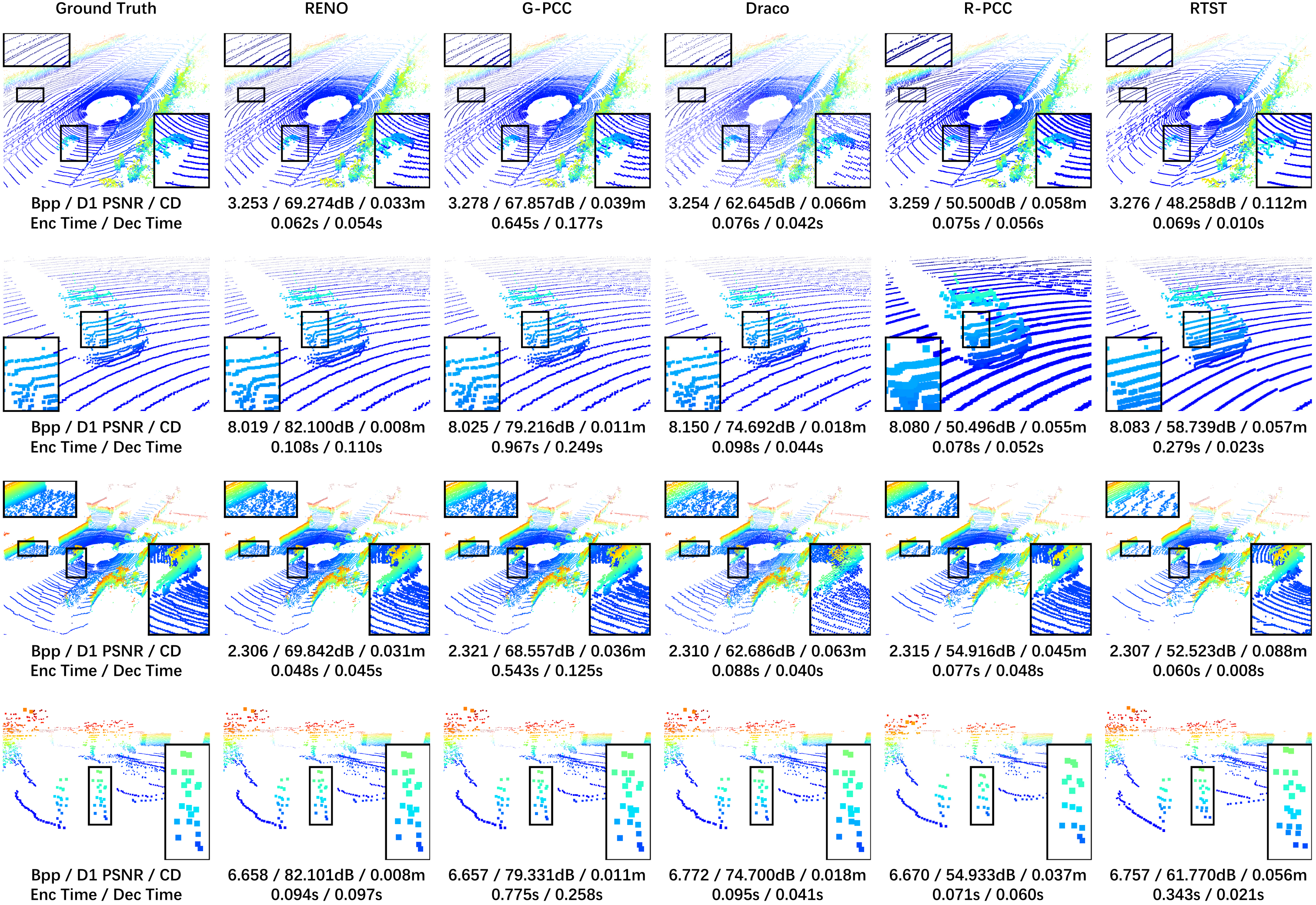}
    \caption{Qualitative visualization. CD refers to Chamfer Distance. The first and third rows indicate the compression results at lower bit rates, while the second and fourth rows indicate compression results at higher bit rates with lower distortion. The color indicates the height above ground, ranging from blue (low) to red (high).}
    \label{fig:subjective_results}
\end{figure*}

\begin{figure}
    \centering
    \includegraphics[width=1.0\linewidth]{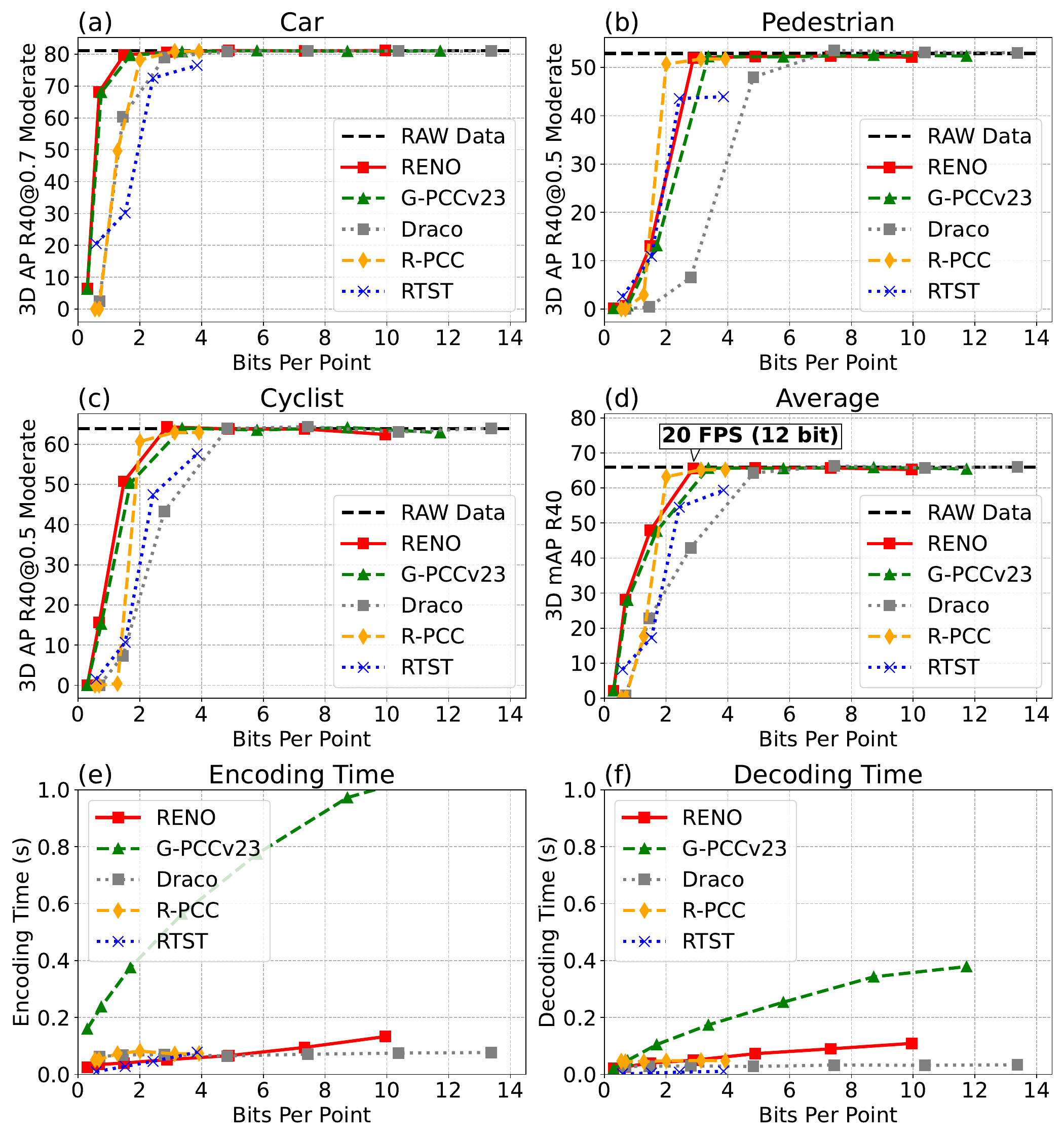}
    \caption{3D object detection on the KITTI dataset.}
    \label{fig:kitti_det}
\end{figure}


\subsubsection{Qualitative Visualization} \Cref{fig:subjective_results} visualizes the reconstructed samples of various methods in both low-bitrate and high-precision scenarios, using two KITTI point clouds as examples. It is evident that our method achieves significantly superior reconstruction quality compared to other real-time methods. For instance, RENO effectively reconstructs the scan strips and vehicle contours at low bit rate conditions. In contrast, Draco appears blurry, and both RTST and R-PCC exhibit a significant lack of scan integrity. At high bit rates, RENO presents an accurate point reconstruction, providing a Chamfer Distance of only 8 mm, which is less than half of Draco's. We can also observe that, despite a substantial bitrate budget, range image-based methods still struggle to reconstruct points accurately, which confirms the significant distortion caused by 2D projection.

\subsubsection{Downstream Task Evaluation} 
Criticism often arises: Why must we compress LiDAR if we can perform vision tasks on front devices using raw captures? One reason is that we can store on-device LiDAR for late (autopilot) simulation purposes. On the other hand, efficient LiDAR can potentially enable device-to-device or vehicle-to-vehicle sharing with rich LiDAR information for collaborative decision-making.
To fulfill it, we must assure the same task accuracy when using a compressed LiDAR sequence as when using uncompressed raw data.


We thus evaluate the performance of a prevalent 3D object detection task using decoded point clouds generated by various real-time LPCCs. We utilize the CenterPoint~\cite{CenterPoint} as the detector, which is trained on raw data, and apply it directly to decoded point clouds. As seen from~\cref{fig:kitti_det} (a)-(d), our approach exhibits substantially superior precision to Draco and exhibits enhanced stability relative to the range image-based baseline (e.g., R-PCC), especially under low bitrate conditions. G-PCC also demonstrates strong detection performance; however, as illustrated in~\cref{fig:kitti_det} (e) and (f), this performance is accompanied by a significant coding latency, rendering it unsuitable for real-time applications. As demonstrated in~\cref{fig:kitti_det} (d), on average, a 12-bit point cloud is sufficient to attain an accuracy close to that of raw LiDAR scan (65.64 mAP vs. 65.97 mAP), and our method can achieve a processing speed of 20 frames per second under 12-bit condition.



\subsection{Ablation Studies}
\label{sec:ablation_study}

\begin{table}[t]
  \centering
  \small
    \caption{Target embedding vs. target convolution. BD-BR gain is against G-PCCv23 in D1 metric.  ``w/o Both'' means the omission of scale-wise feature inheritance. The network inference time for 14 bit KITTI point clouds is reported.}
    \label{tab:ablation_emb}
  \scalebox{0.86}{
      \begin{tabular}{llrrr}
        \toprule
        Module & Implementation & BD-BR (\%) & Inference (s) \\
        \midrule
        w/o Both & MinkowskiEngine~\cite{MinkowskiEngine} & -6.05 & 0.045 \\
        w/o Both & TorchSparse++~\cite{tangandyang2023torchsparse} & -6.06 & 0.024 \\
        \midrule
        w/ TargetConv & MinkowskiEngine~\cite{MinkowskiEngine}    & -8.53 & 0.101 \\
        w/ TargetConv & TorchSparse++~\cite{tangandyang2023torchsparse}      & - & - \\
        \midrule
        w/ TargetEmb & MinkowskiEngine~\cite{MinkowskiEngine}     & -12.24 & 0.071 \\
        \rowcolor[gray]{0.9}
        w/ TargetEmb & TorchSparse++~\cite{tangandyang2023torchsparse}       & -12.26 & 0.035 \\
        \bottomrule
      \end{tabular}
  }

\end{table}

\begin{table}[t]
  \centering
  \small
    \caption{Time consumption for Fast Occupancy Generator (FOG), Neural Network Inference (NN), and Arithmetic Encoding (AE).}
    \label{tab:ablation_stage}
  \scalebox{0.88}{
      \begin{tabular}{lrrrrr}
        \toprule
        & BD-BR (\%) & FOG (s)  & NN (s) & AE (s) & Total (s)\\
        \midrule
        One-Stage & -11.45 & 0.008 & 0.036 & 0.102 & 0.146 \\
        \rowcolor[gray]{0.9}
        Two-Stage & -12.26 & 0.008 & 0.035 & 0.052 & 0.095 \\
        \bottomrule
      \end{tabular}
  }

\end{table}

\textbf{Target Embedding vs. Target Convolution.} In Target Occupancy Predictor (TOP), we employ a target embedding (TargetEmb) to inherit features from the lower scale. Alternatively, we can apply the ``convolution on target coordinates'' operation~\cite{akhtar2024inter,wang2023dynamic}, referred to as target convolution (TargetConv), to transfer the features of the previous scale to the current scale. However, TargetConv is not supported by TorchSparse++ but by MinkowskiEngine platform. 

\Cref{tab:ablation_emb} presents ablation experiments comparing both modules, where we reimplement our method on the MinkowskiEngine~\cite{MinkowskiEngine} library for the support of target convolution operation. As seen, the target embedding significantly outperforms target convolution in both inference speed and rate-distortion performance



\begin{figure}[t]
    \centering
    \includegraphics[width=1.0\linewidth]{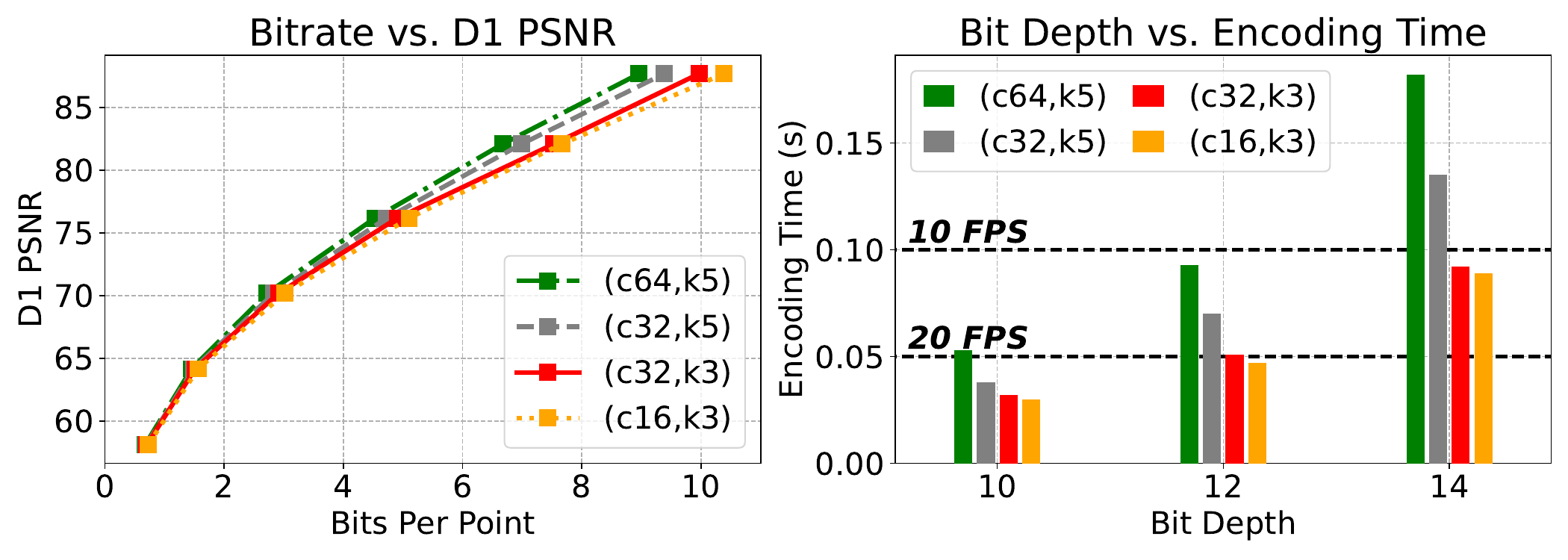}
    \caption{Analysis on channel and kernel size of our neural network. ``(c64, k5)'' denotes the model with 64 channels and a kernel size of 5, with other notations following a similar pattern.}
    \label{fig:ablation_channel_kernel}
\end{figure}

\textbf{Channel and Kernel Size Analysis.} Intuitively, increasing the number of channels and the kernel size contributes to a higher compression ratio, but at the expense of increased inference latency. \Cref{fig:ablation_channel_kernel} shows the results of different channels and kernel sizes. This paper adopts a ``(c32,k3)'' configuration to achieve a balance between performance and latency on our current platform. Nevertheless, with the availability of more powerful hardware, the neural model can be scaled up to achieve better compression ratios.

\textbf{Bitwise Two-Stage Coding.} The bitwise two-stage coding scheme is introduced to encode the occupancy codes in a two-stage manner. To validate its efficacy, a one-stage baseline is established for comparison. The results presented in~\Cref{tab:ablation_stage} demonstrate that the bitwise two-stage coding scheme significantly reduces the latency of arithmetic coding operations, thereby facilitating real-time compression.

Please refer to our supplementary material for more ablation studies and discussions.
\section{Conclution}



This paper presents RENO - a LiDAR coder built upon multiscale sparse representation. At each scale, it compresses sparse occupancy codes to infer voxels' occupancy in a one-shot manner, where the derivation of sparse occupancy codes utilizes fixed-weights sparse convolution without resorting to complex octree construction, and the compression of these codes exploits cross-scale correlations through conditional coding.  RENO offers superior performance at a real-time compression speed, which is attractive for applications like autonomous machinery to save and exchange instantly captured LiDAR data for better decision-making.


\section{Acknowledgement}

This work was supported in part by the Key Project of Jiangsu Science and Technology Department under Grant BK20243038, and in part by the Key Project of the National Natural Science Foundation of China under Grant 62431011. The authors would like to express their sincere gratitude to the Interdisciplinary Research Center for Future Intelligent Chips (Chip-X) and Yachen Foundation for their invaluable support.

{
    \small
    \bibliographystyle{ieeenat_fullname}
    \bibliography{main}
}

\clearpage
\setcounter{page}{1}
\maketitlesupplementary

\section{Overview}
This appendix provides supplementary comparisons and discussions to complement the manuscript. \Cref{sec:supp_learning_based_comparison} compares our approach with learning-based methods, while \cref{sec:supp_ablation} presents additional ablation studies. Implementation details are discussed in \cref{sec:supp_implementation}, and more visualizations are provided in \cref{sec:supp_subjective}. All experiments in this appendix are conducted on an Intel Xeon Platinum 8352V CPU and one RTX 4090 GPU.


\section{Comparison with Learning-based Methods}
\label{sec:supp_learning_based_comparison}

In this section, we provide a detailed comparison of RENO with representative learning-based methods: EHEM~\cite{EHEM} and Unicorn~\cite{Unicorn_Part1}. EHEM is recognized as the most prominent and high-performing model among tree-based approaches~\cite{EHEM,ECM_OPCC,OctAttention,OctSqueeze}, while Unicorn exemplifies the latest and most representative sparse tensor-based models~\cite{Unicorn_Part1,xue2022efficient,SparsePCGC}. \Cref{tab:neural_model_compare} shows the rate distortion performance of different methods, where RENO-L represents the large version of RENO, which increases the number of channels from 32 to 128 and the kernel size from 3 to 5. \Cref{tab:neural_model_time} provides detailed encoding time for various methods applied to 12-bit and 14-bit point clouds, where the first frame in SemanticKITTI sequence 11 is used for test.

\begin{itemize}
    \item It can be observed that EHEM demonstrates the highest rate-distortion performance (note that the BD-BR of EHEM is directly cited from their paper and is presented for reference only); however, this comes at the cost of substantial time consumption. Its preprocessing stage demands significantly more time to construct the tree and prepare contextual information, ultimately undermining its real-time applicability. 
    
    \item In contrast, Unicorn performs inference directly within the sparse tensor domain, effectively minimizing preprocessing overhead. However, its reliance on the upsampling-based inference framework results in a substantially prolonged neural network computation time. While the inference speed of the network can be enhanced using 1-stage inference, Unicorn\textsuperscript{1} fails to achieve substantial speed improvements (as it still necessitates the introduction of a large number of voxels via deconvolution) and results in a significant degradation in performance.
\end{itemize}

\begin{figure}[t]
    \centering
    \includegraphics[width=1.0\linewidth]{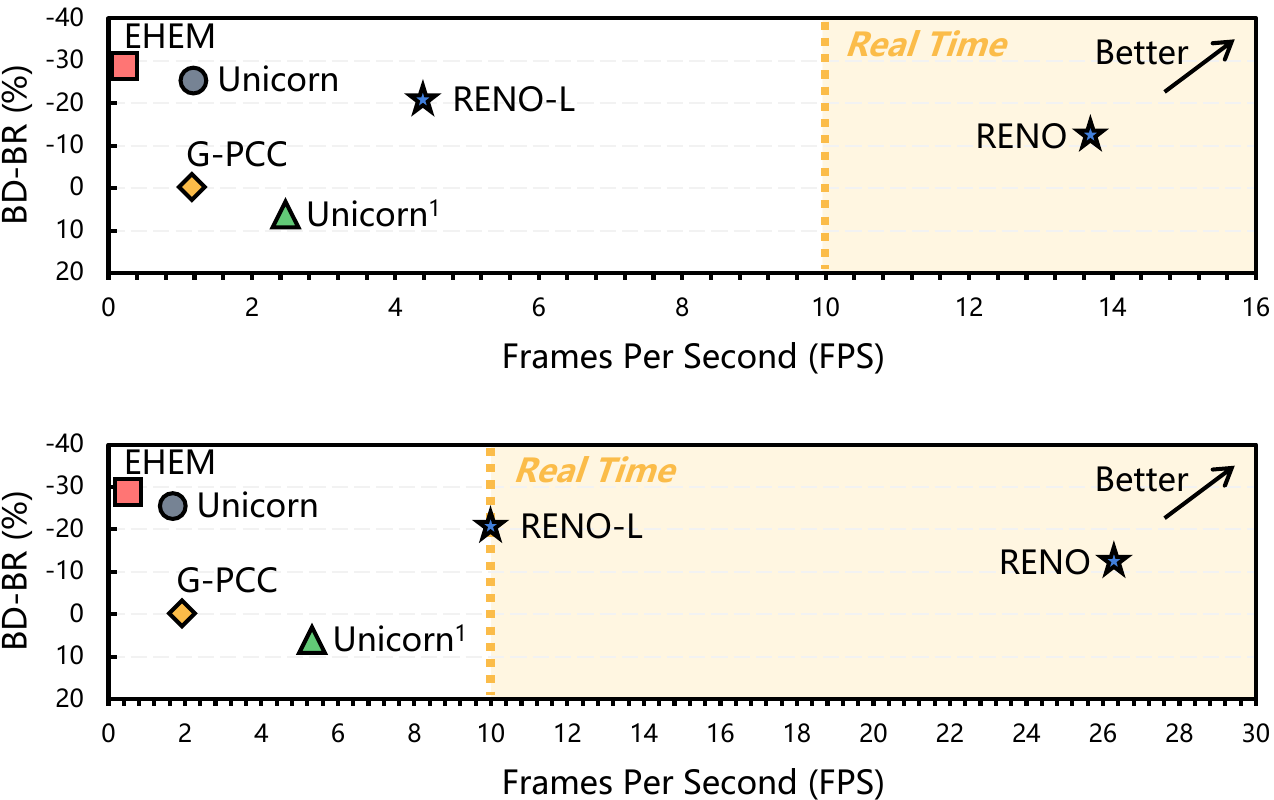}
    \caption{Comparison with the learning-based methods EHEM and Unicorn. The encoding speeds are reported separately for 14-bit precision (top) and 12-bit precision (bottom). Unicorn\textsuperscript{1} refers to the one-stage configuration of the Unicorn model. G-PCC serves as the anchor.}
    \label{fig:supp_neural_time}
\end{figure}

{
\setlength{\tabcolsep}{0.5em} 
\begin{table}[h]
  \centering
  \small
  \caption{Rate-distortion comparison with learning-based methods on SemanticKITTI. BD-BR refers to the BD-BR gain over G-PCCv23 in D1 metric. Unicorn\textsuperscript{1} denotes the one-stage configuration of the Unicorn model.}
  \scalebox{0.84}{
      \begin{tabular}{cccccc}
        \toprule
        Method & EHEM & Unicorn & Unicorn\textsuperscript{1} & RENO-L & RENO \\
        \midrule
        BD-BR (\%) & -28.89 & -25.62 & +6.56 & -20.63 & -12.47 \\
        \bottomrule
      \end{tabular}
  }
    \label{tab:neural_model_compare}
\end{table}
}

{
\setlength{\tabcolsep}{0.53em} 
\begin{table}[h]
  \centering
  \small
  \caption{Encoding time comparison. Preprocessing (Prep), neural network inference (NN), and arithmetic encoding (AE) time are independently reported for detailed comparison. $D$ refers to the bit depth of reconstructed point clouds. For sparse tensor-based methods (e.g., Unicorn, Unicorn\textsuperscript{1}, RENO-L, and RENO), the time spent on dyadic downscaling is reported as preprocessing.}
  \scalebox{0.8}{
      \begin{tabular}{lrrrrrrrrr}
        \toprule
        \multirow{2}[2]{*}{Method} & \multicolumn{4}{c}{$D$=12 (s)} & \multicolumn{4}{c}{$D$=14 (s)} \\
        \cmidrule(lr){2-5} \cmidrule(lr){6-9}
        & Prep & NN & AE & \textbf{Total} & Prep & NN & AE & \textbf{Total} \\
        \midrule
        EHEM    & 0.463       & 0.427 & 0.297 & \textbf{1.187}      & 1.796 & 1.385 & 1.233 & \textbf{4.414} \\
        Unicorn & 0.003 & 0.565 & 0.030 & \textbf{0.598}         & 0.003 & 0.772 & 0.074 & \textbf{0.849} \\
        Unicorn\textsuperscript{1} & 0.003 & 0.163 & 0.021 & \textbf{0.187}         & 0.003 & 0.336 & 0.063 & \textbf{0.402} \\
        \rowcolor[gray]{0.9}
        RENO-L & 0.006     & 0.080 & 0.014 & \textbf{0.100}     & 0.007 & 0.184 & 0.037 & \textbf{0.228} \\
        \rowcolor[gray]{0.9}
        RENO   & 0.006 & 0.018 & 0.014 & \textbf{0.038}    & 0.007 & 0.028 & 0.038 & \textbf{0.073} \\
        \bottomrule
      \end{tabular}
  }
    \label{tab:neural_model_time}
\end{table}
}


\section{Additional Ablation Study}
\label{sec:supp_ablation}

\textbf{CPU vs. GPU Arithmetic Coding.} This paper employs a bitwise two-stage coding strategy to  accelerate the arithmetic coding process, leveraging the Torchac\footnote{\href{https://github.com/fab-jul/torchac}{https://github.com/fab-jul/torchac}} library, which operates on the CPU. An alternative acceleration strategy involves dividing symbols into packets for parallel coding directly on the GPU, an approach implemented in GPUAC\footnote{\href{https://github.com/zb12138/GPUAC}{https://github.com/zb12138/GPUAC}}. Here, we performed ablation experiments for both methods, and the results are shown in \cref{tab:cpu_gpu_ac}. It should be noted that an independent arithmetic encoding process is conducted for each scale, and the reported time represents the accumulated arithmetic encoding time across all scales. The default packet size of 8192 is used when implementing GPUAC. As observed, GPUAC provides a slight speedup over the naive one-stage approach only at higher bitrates ($D$=14); however, this performance improvement is considerably inferior compared to the proposed bitwise decomposition strategy. At lower bitrates ($D$=12), GPUAC performs significantly slower, as the number of symbols is relatively small, and the CPU is sufficiently capable of processing the data with low latency.

\begin{table}[t]
  \centering
  \small
  \caption{Comparison of arithmetic coding implementations. The first frame in the SemanticKITTI sequence 11 is used for testing. Encoding time for different bit depths $D$ are reported.}
  \scalebox{0.85}{
      \begin{tabular}{llrrrrrrrr}
        \toprule
        \multirow{2}[2]{*}{Module} & \multirow{2}[2]{*}{Library} & \multicolumn{2}{c}{$D$=12} & \multicolumn{2}{c}{$D$=14} \\
        \cmidrule(lr){3-4} \cmidrule(lr){5-6}
        & & Time (s) & Bitrate & Time (s) & Bitrate \\
        \midrule
        One-Stage & GPUAC & 0.065 & 2.24       & 0.096 & 6.56 \\
        Two-Stage & GPUAC & 0.064 & 2.25       & 0.093 & 6.55 \\
        \midrule
        One-Stage & Torchac & 0.017 & 2.27     & 0.102 & 6.58 \\
        \rowcolor[gray]{0.9}
        Two-Stage & Torchac & 0.014 & 2.28     & 0.038 & 6.57 \\
        \bottomrule
      \end{tabular}
  }
    \label{tab:cpu_gpu_ac}
\end{table}



\section{Implementation Details}
\label{sec:supp_implementation}

\subsection{Detailed Network Structure}

\begin{figure}[t]
    \centering
    \includegraphics[width=0.6\linewidth]{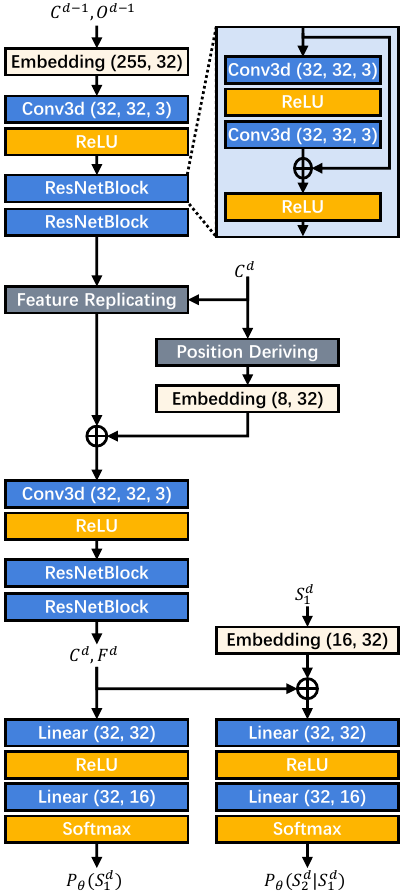}
    \caption{Detailed network structure of the devised Target Occupancy Predictor (TOP). ``Conv3d'' represents 3D sparse convolution, with parameters defined as (input channels, output channels, kernel size); ``Embedding'' maps discrete variables according to (dictionary size, vector dimension); ``Linear'' applies an affine linear transformation with parameters (input channels, output channels); ``$\oplus$'' denotes element-wise addition.}
    \label{fig:supp_top}
\end{figure}

\Cref{fig:supp_top} illustrates the neural network architecture and detailed parameters employed by RENO. The model parameters are shared across all point cloud scales.

\subsection{Configuration of Comparative Methods}

\textbf{G-PCCv23}\footnote{\href{https://github.com/MPEGGroup/mpeg-pcc-tmc13}{https://github.com/MPEGGroup/mpeg-pcc-tmc13}} is employed following the Common Test Condition (CTC) recommended by the MPEG committee. For KITTI point clouds, the coordinates are scaled by dividing by 0.001, and the $posQ$ parameter is adjusted to achieve different bitrates. After decoding, the reconstructed point clouds are rescaled to the original coordinate system by multiplying by 0.001. For Ford point clouds, as they are already quantized to 1 mm precision, the original coordinate system is retained. 

\textbf{DracoPy}\footnote{\href{https://github.com/seung-lab/DracoPy}{https://github.com/seung-lab/DracoPy}} , utilizing Google's Draco version 1.5.2, is employed for compression in this study. Despite its longevity, Draco remains a benchmark for real-time scenarios~\cite{liang2024fumos}. The compression level is maintained at its default setting of 1, while the quantization bits are adjusted to achieve varying compression ratios.

\textbf{R-PCC}\footnote{\href{https://github.com/StevenWang30/R-PCC}{https://github.com/StevenWang30/R-PCC}} is considered as a baseline for range image-based compression. The BZip2 compressor is employed for the optimal compression rate, and varying compression ratios are achieved by adjusting the $accuracy$ parameter. 

\textbf{RTST}\footnote{\href{https://github.com/horizon-research/Real-Time-Spatio-Temporal-LiDAR-Point-Cloud-Compression}{https://github.com/horizon-research/Real-Time-Spatio-Temporal-LiDAR-Point-Cloud-Compression}} is employed as an additional range image-based compression approach. Single-frame compression mode is adopted, and the horizontal and vertical degree granularities are systematically varied to attain different compression ratios. Note that it is currently supports only the KITTI dataset.

\begin{figure}[t]
  \centering
  \includegraphics[width=1.0\linewidth]{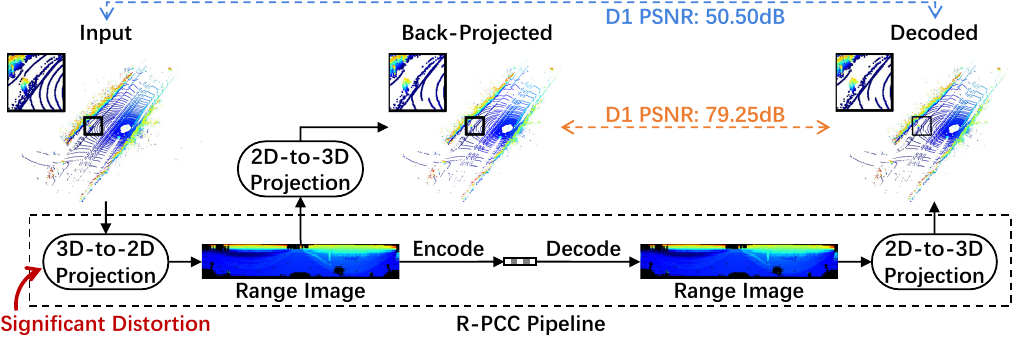}
   \caption{Ours (\textcolor{NavyBlue}{blue line}) vs. R-PCC (\textcolor{orange}{orange line}) PSNR evaluations. A point cloud from the KITTI dataset is utilized as an exemplar. The setting of PSNR peak value is consistent at 59.70.}
   \label{fig:RPCC_PSNR}
\end{figure}

\textbf{Remark.} We observe a notable discrepancy between our experimental results and those reported in the original R-PCC paper~\cite{R-PCC}, which reported markedly superior performance of R-PCC over G-PCC. This divergence primarily stems from methodological differences in PSNR computation. Specifically, the original implementation of R-PCC (as verified through their GitHub repository) employs back-projected geometry rather than original point cloud data as ground truth, thereby circumventing projection-induced distortions illustrated in~\cref{fig:RPCC_PSNR}. The projection distortion is confirmed in both prior studies~\cite{wu2021detailed} and our experiments.  In contrast, our calculation is more reasonable and aligns with prior arts.

\subsection{Metrics}
\textbf{PSNR.} The MPEG PCC quality measurement software version 0.13.4 is used to report PSNR values. The peak values are set to 59.70 for KITTI point clouds and 30,000 for Ford point clouds, following conventional practices~\cite{EHEM,OctAttention,OctSqueeze}.

\textbf{Chamfer Distance.} The Chamfer Distance in the Fig. 1 of the main manuscript serves as an auxiliary description of point-level distortion. This metric is computed using the following mathematical formulation, implemented according to the Point Cloud Utils\footnote{\href{https://fwilliams.info/point-cloud-utils/sections/shape_metrics/}{https://fwilliams.info/point-cloud-utils/sections/shape\_metrics/}}:
\begin{align}
    \mathrm{chamfer}\left(P_1, P_2 \right) = & \ \frac{1}{2n}\sum_{i=1}^{n}| x_i - \mathrm{NN}\left( x_i, P_2 \right)| \nonumber\\
    & + \frac{1}{2m}\sum_{j=1}^{m}| x_j - \mathrm{NN}\left( x_j, P_1 \right)|
\end{align} 
where $P_1 = \left\{ x_i \right\}_{i=1}^{n}$ and $P_2 = \left\{ x_j \right\}_{j=1}^{m}$ refer to two point cloud samples; $\mathrm{NN} \left(x, P\right) = \mathrm{argmin}_{x' \in P} \left \| x - x’ \right \| $ is the nearest neighbor function.

\section{More Subjective Results}
\label{sec:supp_subjective}

\begin{figure*}
    \centering
    \includegraphics[width=1.0\textwidth]{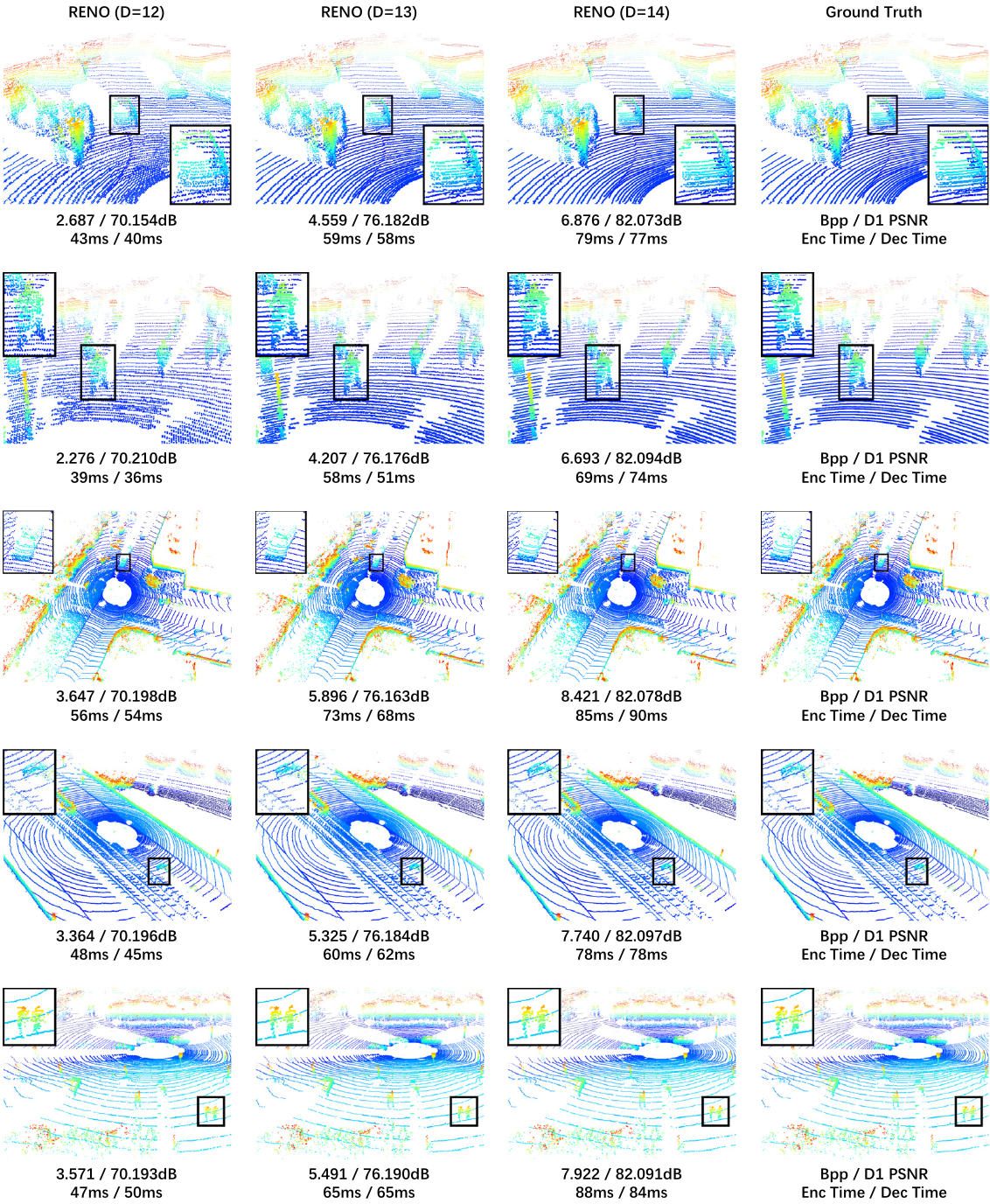}
    \caption{Visualization of compression results for KITTI point cloud. The reconstruction results of RENO under bit depths ($D$) of 12, 13, and 14 are presented.}
    \label{fig:supp_visualization}
\end{figure*}

More visualization results are provided in~\cref{fig:supp_visualization}. As seen from the leftmost column ($D$=12), RENO reconstructs point cloud samples with shapes sufficient to distinguish object categories at around 50 ms latency, differing mainly in fine details. For 14-bit precision ($D$=14), it achieves real-time reconstruction with details closely matching the originals.

\end{document}